\documentclass[conference]{IEEEtran}
\usepackage[numbers,sort]{natbib}
\IEEEoverridecommandlockouts
\usepackage{graphicx} 
\usepackage{float}
\usepackage{amsmath,amssymb,amsfonts}
\usepackage{algorithmic}
\usepackage{graphicx}
\usepackage{textcomp}
\usepackage{xcolor}
\usepackage{hyperref}

\hypersetup{
    colorlinks=true,
    linkcolor=black,
    urlcolor=black,
    citecolor=black,
}

\def\BibTeX{{\rm B\kern-.05em{\sc i\kern-.025em b}\kern-.08em
    T\kern-.1667em\lower.7ex\hbox{E}\kern-.125emX}}
\begin{document}

\title{Real Time Emotion Analysis Using Deep Learning for Education, Entertainment, and Beyond}

\author{\IEEEauthorblockN{Abhilash Khuntia}
\IEEEauthorblockA{\textit{M.Tech CSE - MT23007} \\
\textit{IIITD}\\
abhilash23007@iiitd.ac.in}
\and
\IEEEauthorblockN{Shubham Kale}
\IEEEauthorblockA{\textit{M.Tech CSE - MT23094} \\
\textit{IIITD}\\
shubham23094@iiitd.ac.in}
}

\maketitle

\begin{abstract}
The significance of emotion detection is increasing in education, entertainment, and various other domains. We are developing a system that can identify and transform facial expressions into emojis to provide immediate feedback.The project consists of two components. Initially, we will employ sophisticated image processing techniques and neural networks to construct a deep learning model capable of precisely categorising facial expressions. Next, we will develop a basic application that records live video using the camera on your device. The app will utilise a sophisticated model to promptly analyse facial expressions and promptly exhibit corresponding emojis.Our objective is to develop a dynamic tool that integrates deep learning and real-time video processing for the purposes of online education, virtual events, gaming, and enhancing user experience. This tool enhances interactions and introduces novel emotional intelligence technologies.
\end{abstract}

\section{Motivation}
Emojis are great way to express our feelings in the modern world of communication. Many social media sites like Facebook, Instagram, Twitter use this feature. However, manually selecting emojis can be time-consuming and often inaccurate. Leveraging facial detection technology to predict emojis has the potential to transform this process, offering users a seamless and intuitive way to express themselves in digital conversations.\par
This endeavor embarks on the development of a cutting-edge real-time Facial Expression Recognition (FER) system, seamlessly integrating the prowess of OpenCV, TensorFlow, and NumPy \cite{abadi2016tensorflow, pedregosa2011scikit, lecun1998gradient}. OpenCV stands as the cornerstone, furnishing a stalwart framework for image and video analysis. TensorFlow, on the other hand, emerges as an indispensable asset, furnishing an expansive platform for machine learning prowess. Complementing these formidable tools, NumPy emerges as the quintessential companion, demonstrating unparalleled efficiency in handling vast, multi-dimensional arrays. Together, these technologies converge to forge an unparalleled solution, poised to redefine the landscape of facial expression recognition in real-time.\par

\section{Literature Review}
CNN-based approach had the following model and workflow: training the model on the FER2013 dataset (Facial emotion recognition dataset), which is a very famous dataset, followed by gathering the CNN weights, and testing on images/videos captured by webcams/cameras and then mapping the result to appropriate emojis. The face detection step is carried forward with the help of Haar cascade detection from the OpenCV library \cite{viola2001rapid}. The CNN model developed is trained to recognize human facial features and ignore other shapes, objects, or landscapes.\par
\textbf{H. Guo and J. Chen} in their paper \cite{guo2019facial} proposed to use a combination of CNN and LSTM. The proposed network has 4 layers - input layer, CNN Feature sampling layer, LSTM feature learning layer and finally the output layer which here is the SVM feature classification layer. In the CNN layer the authors have proposed to use a ResNet model. The LSTM layer which is responsible for learning the features uses the loop of RNNs (Recurrent Neural Networks) and information memory of LSTM \cite{hochreiter1997long}. Then, in the end from the features learnt and information gathered, the SVM layer classifies the input into various classes \cite{burges1998tutorial}.\par
Y. Zhou et al. proposed to use ExpressionNet for CNN trained by using Caffe, a deep learning framework \cite{zhou2020expressionnet}. The ExpressionNet model comprises 9 modules of ReduceV2. Each of ReduceV2 module is followed by a BN layer and ReLU activation function. A ReduceV2 module is used to split the convolution layer into a dimension reduction layer and a sampling layer and each sampling layer has a ReLU active layer.\par
A single standalone CNN model is the basis of the proposed architecture in \cite{li2019facial}, which is implemented on a real-time Intelligent System for Sentiment Recognition. The system uses transfer learning to validate accuracy and combines several tasks into a single blended step, including face detection, sentiment classification, and the provision of a live list of probabilistic labels from a webcam feed.\par

\section{Dataset Details}
The FER2013 dataset contains 48x48 pixel grey images of faces representing different emotions \cite{goodfellow2016deep}.
\begin{figure}[h!] 
    \centering
    \includegraphics[width=0.80\linewidth]{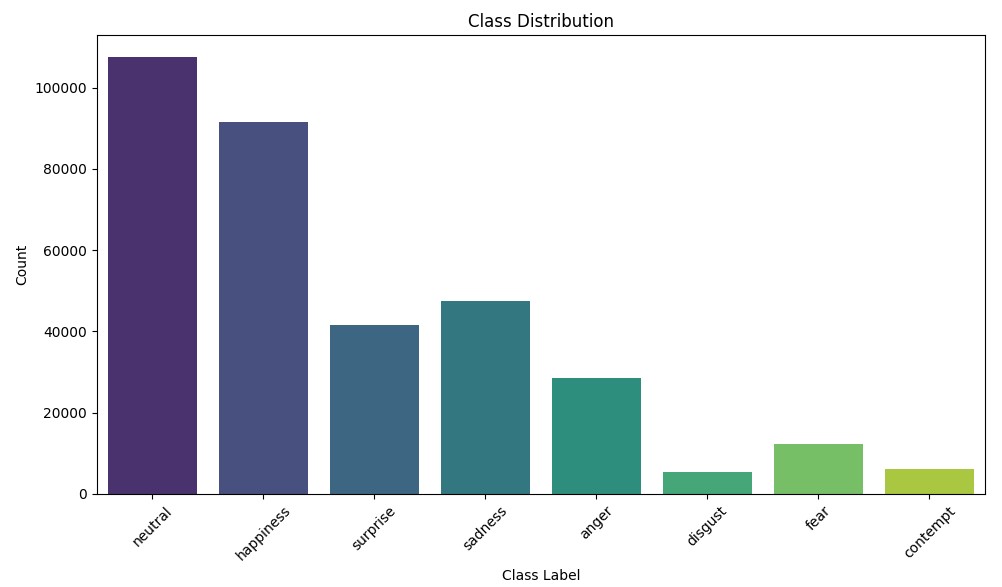}
    \caption{Data Imbalance}
    \label{fig:imbalence}
\end{figure}
The faces have been automatically registered so that the face is more or less centred and occupies about the same amount of space in each image. There are 7 labels denoting each emotion - 0=Angry, 1=Disgust, 2=Fear, 3=Happy, 4=Sad, 5=Surprise, 6=Neutral. The dataset is already split into train and test, having 28,709 examples and 3,546 examples respectively. The Dataset also have imbalance which can be seen in the figure \ref{fig:imbalence} \cite{mollahosseini2016affectnet}.\par

\section{Proposed Architecture}
For this project we have planned to use 4 approaches -
1. CNN
2. LSTM
3. ResNet
4. QDA + PCA \cite{shlens2014tutorial}

\begin{figure}[h!]
    \centering
    \includegraphics[width=1\linewidth]{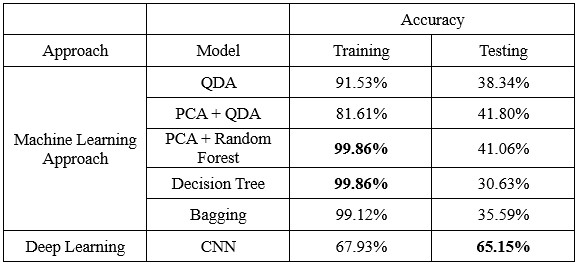}  
    \caption{Accuracy Using Different Approaches.}
    \label{fig:Accuracy}
\end{figure}

From the below given table, we will be moving forward with CNN model \cite{krizhevsky2012imagenet}. Also one of the main reasons of CHOOSING CNN over any traditional ML techniques, apart from accuracy is that in Machine learning approach we have to do manual feature extraction from images. Whereas, CNNs during training, features are automatically learned from raw data. CNNs use a sequence of convolutional layers to learn hierarchical representations of features from the input data, as opposed to manually creating and extracting features \cite{simonyan2014very}. CNNs are incredibly efficient at tasks like object detection, facial recognition, and picture classification because of these learnt characteristics that allow them to catch patterns and spatial hierarchies in the data \cite{ren2015faster}.

After finalizing our modeling technique, our next goal is to improve the above testing accuracy.

\subsection{Initial CNN Architecture}

The initial CNN architecture that we built gave us training accuracy of \textbf{67.93\%} and a testing accuracy of about \textbf{65.15\%}. Below diagram represents the architecture.

\begin{figure}[h!]
    \centering
    \includegraphics[width=0.90\linewidth]{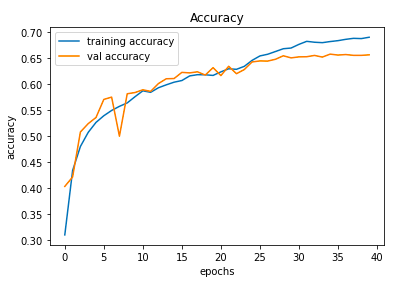}
    \caption{Accuracy v/s Epochs for Training and Validation sets}
    \label{fig:accuracy_epochs}
\end{figure}

\begin{figure}[h!]
    \centering
    \includegraphics[width=0.90\linewidth]{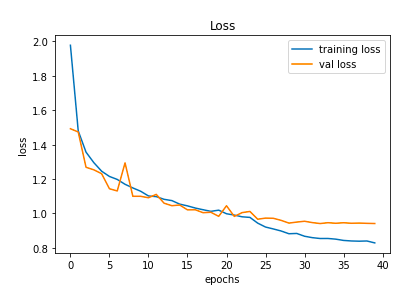}
    \caption{Loss v/s Epochs for Training and Validation sets}
    \label{fig:loss_epochs}
\end{figure}\par

The plots shown in Figure \ref{fig:accuracy_epochs} shows the variation of training accuracy and validation accuracy versus number of epochs. On the other hand in Figure \ref{fig:loss_epochs} shows the variation of training loss and validation loss versus number of epochs.\par

\subsection{Initial ResNet18 Architecture}
\begin{figure}[h!]
    \centering
    \includegraphics[width=0.90\linewidth]{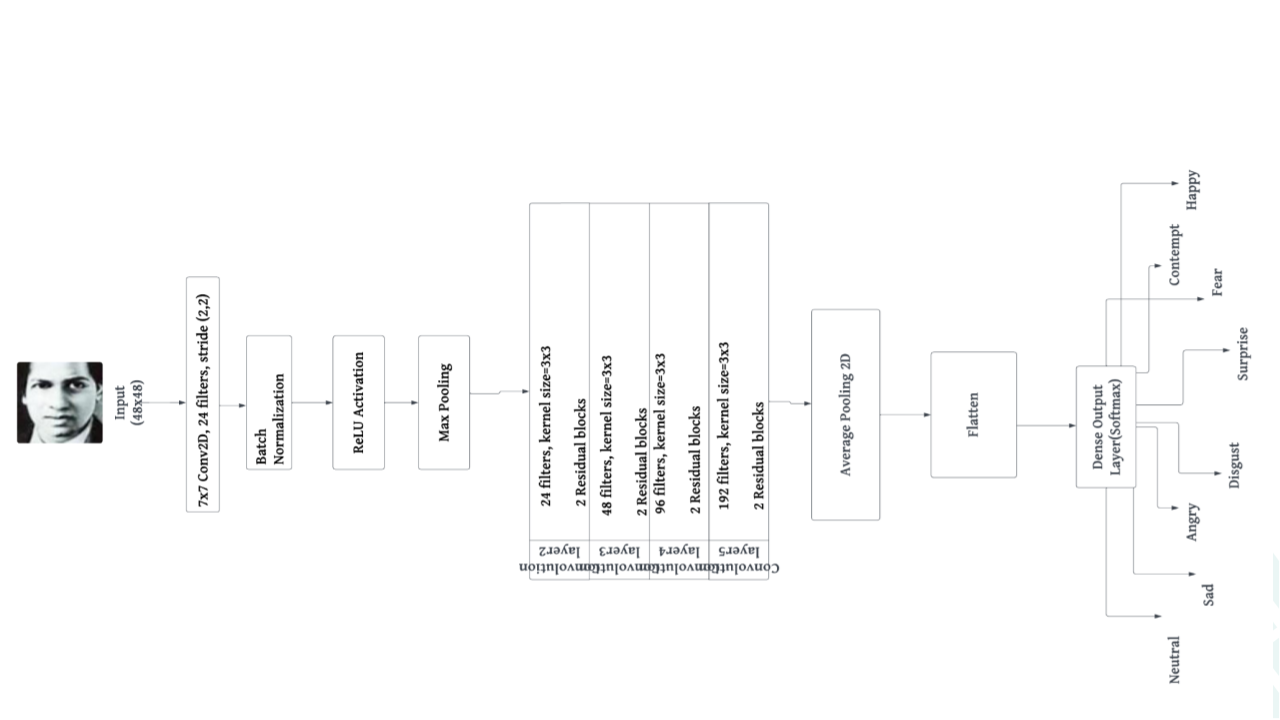}
    \caption{ResNet18 Initial architecture}
    \label{fig:initial_architecture}
\end{figure}

Here we decided to use ResNet18 model of CNN architecture where initially we use some data augmentation techniques in which we use random\_eraser function to randomly erase some part of the given image from our dataset so that we can have data variability \cite{he2016deep}. Also we have considered a different dataset for this approach in which we have 8 emotions named 'neutral', 'happiness', 'surprise', 'sadness', 'anger', 'disgust', 'fear', 'contempt'. We also decided to use the FER2013+ dataset instead of the FER2013 dataset due to numerous labeling errors and some images that don't accurately represent human facial expressions. The FER2013+ dataset, re-annotated by Microsoft (available as \href{https://github.com/microsoft/FERPlus/blob/master/fer2013new.csv}{\textbf{FER+}} on GitHub, provides improved label annotations for the Emotion FER dataset \cite{yin20063d}. Our CNN model is a custom ResNet18 architecture built from scratch using Keras and the architecture diagram is shown in figure \ref{fig:initial_architecture}.

\begin{figure}[h!]
    \centering
    \includegraphics[width=0.90\linewidth]{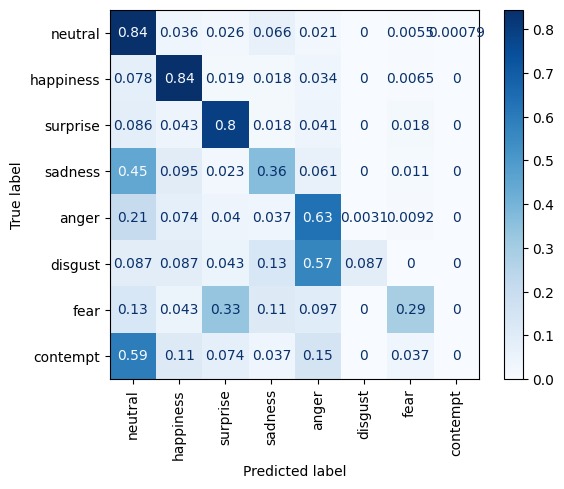}
    \caption{Confusion Matrix}
    \label{fig:cnf_matrix}
\end{figure}

\begin{figure}[h!]
    \centering
    \includegraphics[width=0.90\linewidth, height=5cm]{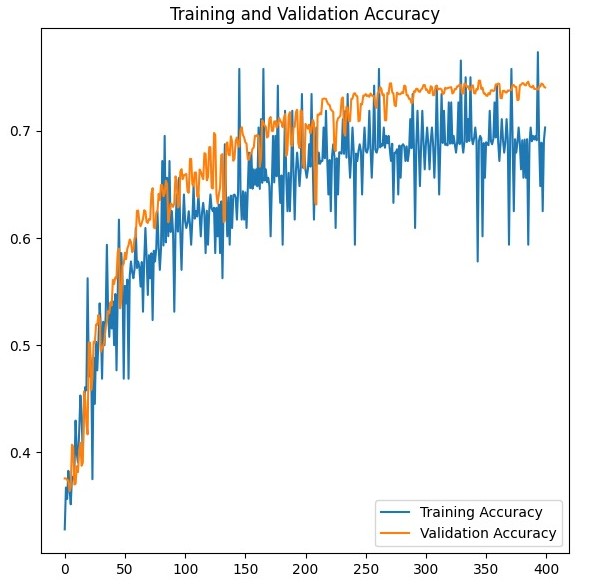}
    \caption{Training And Validation Accuracy}
    \label{fig:train_val_accuracy}
\end{figure}

\begin{figure}[h!]
    \centering
     \includegraphics[width=0.80\linewidth, height=5cm]{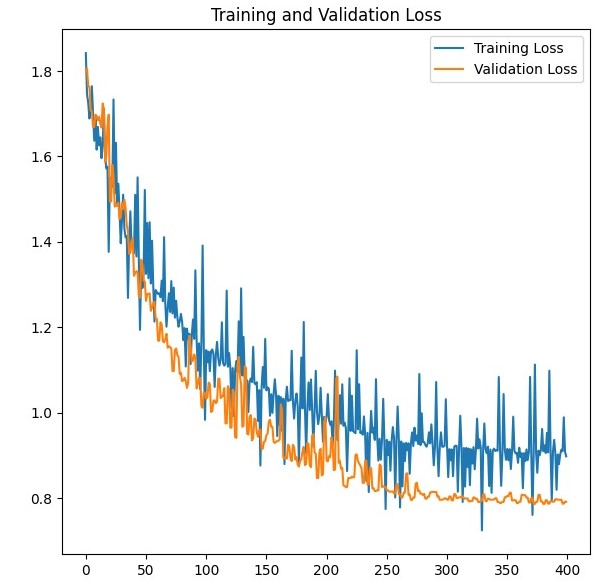}
    \caption{Training and Validation Loss}
    \label{fig:train_val_loss}
\end{figure}

\subsection{Improved ResNet18 Architecture}

\begin{figure}[h!]
    \centering
    \includegraphics[width=0.90\linewidth]{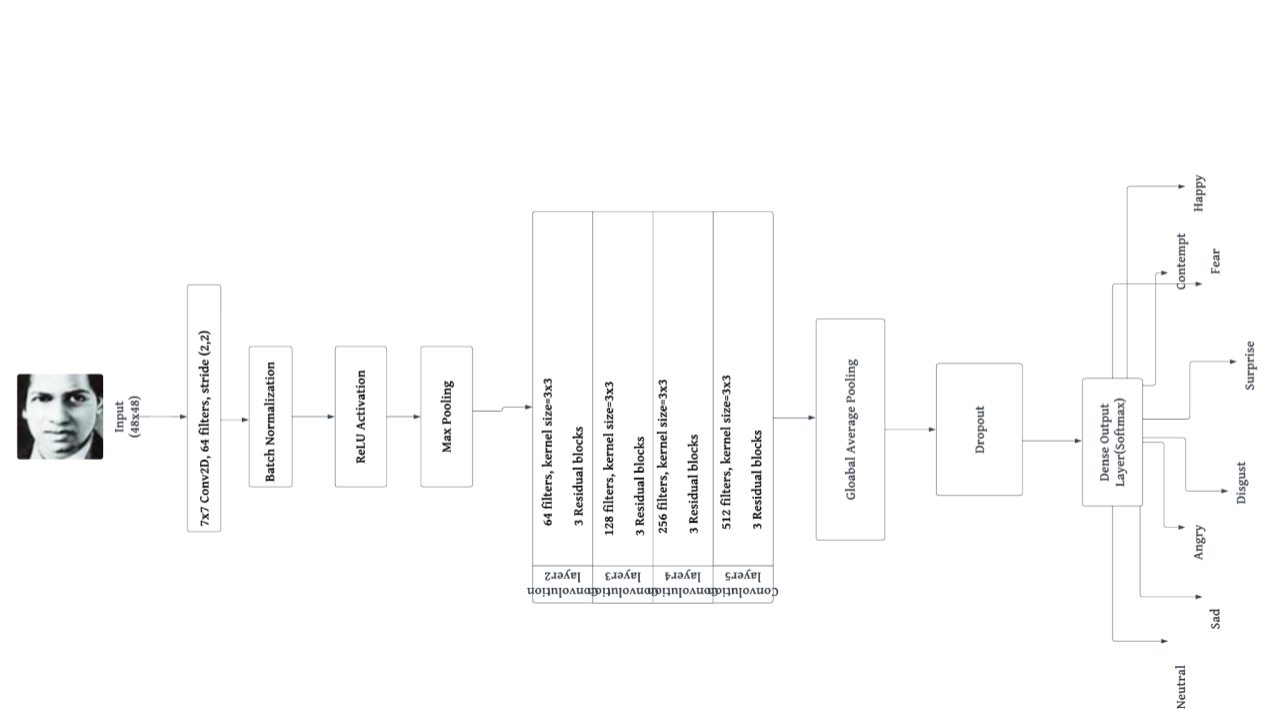}
        \caption{ResNet18 Final architecture}
    \label{fig:final_architecture}
\end{figure}

Here we have changed the initial architecture of the ResNet18 model and made it a more dense architecture by adding more layers which also increased the number of model parameters. We increased the filter size for each convolutional layer. This parameter is the number of filters or channels in the convolutional layers within the residual block. Each filter detects different features in the input data. Also we added Kernel\_Regularizer=l2(weight\_decay) which adds a regularization term to the weight of the convolutional layer. It further penalizes the larger weights in the network to prevent overfitting. The architecture diagram is shown in figure \ref{fig:final_architecture}.

\begin{figure}[h!]
    \centering
    \includegraphics[width=1\linewidth]{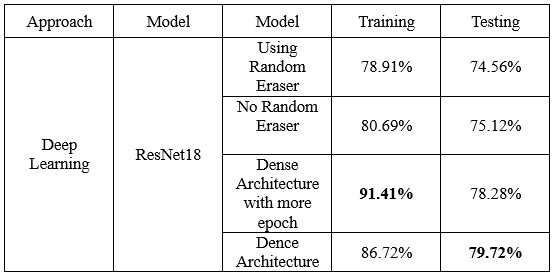}  
    \caption{Accuracy Table with Improved Architecture of ResNet18 Model.}
    \label{fig:Accuracy-table_improved}
\end{figure}

The table shown in figure \ref{fig:Accuracy-table_improved} shows accuracy of different resnet18 improved architectures.

\begin{figure}[h!]
    \centering
    \includegraphics[width=0.70\linewidth]{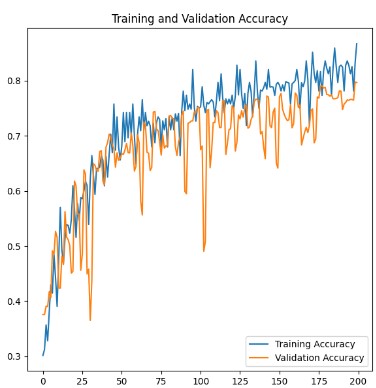}
    \caption{Accuracy v/s Epochs for Training and Validation sets}
    \label{fig:accuracy_epochs_val}
\end{figure}

\begin{figure}[h!]
    \centering
    \includegraphics[width=0.70\linewidth]{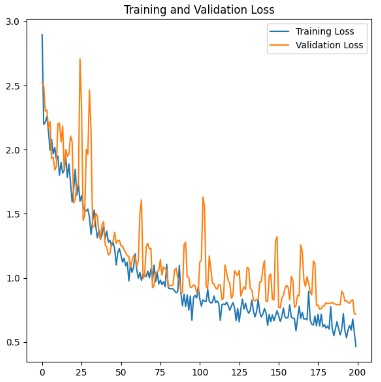}
    \caption{Loss v/s Epochs for Training and Validation sets}
    \label{fig:loss_epochs_val}
\end{figure}

\begin{figure}[h!]
    \centering
    \includegraphics[width=0.80\linewidth]{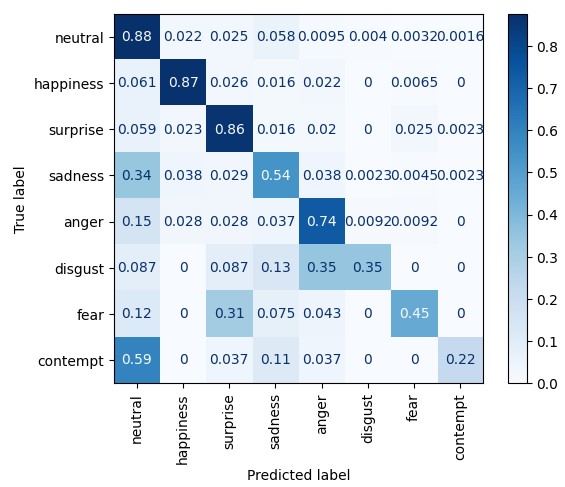}
    \caption{Confusion Matrix of Improved ResNet18}
    \label{fig:cnf_matrix_improved}
\end{figure}
\par

\section{GUI Prototype}

\subsection{Initial Prototype}
This prototype exemplifies the essence of our emotion prediction application, which harnesses the power of our meticulously trained model to discern emotions from facial features captured by webcams or cameras embedded in a multitude of devices, spanning from mobile phones to laptops and beyond.

At its core, this prototype embodies the fusion of cutting-edge technology and human-centric design, aiming to deliver a seamless and intuitive user experience. By leveraging advanced machine learning techniques, our trained model has acquired the ability to analyze facial expressions in real-time, allowing users to interact with the application effortlessly.\par 
Within our GUI, the integration of the Haar Cascade FrontalFace Classifier with OpenCV forms the foundational backbone for extracting crucial facial features from images. This dynamic duo collaborates seamlessly to identify and isolate facial regions within the input images, employing a cascade of classifiers to detect frontal faces accurately.

Once these facial features are extracted, they are seamlessly passed on to our trained model, poised at the heart of our emotion prediction system. Trained meticulously on extensive datasets, our model possesses the proficiency to decipher the subtle nuances of human emotions encoded within these facial features.

Utilizing sophisticated algorithms and neural network architectures, our model meticulously analyzes the extracted facial features, discerning intricate patterns and cues indicative of various emotional states. With each analysis, our model leverages its learned knowledge to predict the most appropriate emotion label corresponding to the input facial expression.

This symbiotic relationship between the Haar Cascade FrontalFace Classifier, OpenCV, and our trained model forms the crux of our GUI's functionality. Together, they seamlessly orchestrate the process of emotion prediction, providing users with insightful and accurate assessments of emotional states encapsulated within the input images.\par
The basic steps involved here are -
\begin{enumerate}
    \item \textbf{Face Detection using Haar Cascade:} Use a pre-trained Haar cascade frontalface classifier to detect faces in the input image. This will provide bounding boxes around the detected faces.

    \item \textbf{Extract Face Regions of Interest (ROIs):} For each detected face, extract the region of interest (ROI) from the input image.

    \item \textbf{Preprocess Face ROIs:} Preprocess the extracted face ROIs. This may involve resizing the ROIs to a fixed size, converting them to grayscale, and normalizing pixel values.

    \item \textbf{Emotion Prediction using ResNet18:} Use a pre-trained ResNet18 model (or any other suitable model) to predict the emotions from the preprocessed face ROIs. The ResNet18 model is trained on FER2013 dataset(with annotatins using FER+ dataset from Microsoft) containing facial images labeled with corresponding emotions (e.g., happy, neutral, angry, sad, disgust, fear, surprise, contempt). Pass the preprocessed face ROIs through the ResNet18 model to obtain predictions for each emotion class.

    \item \textbf{Post-processing Predictions:} The model provides numerical labels between 0-7, then we map it appropriate label - happy, neutral, angry, sad, disgust, fear, surprise, contempt. If, model was not able to classify into one of the above mentioned classes then it outputs "unknown".

    \item \textbf{Display or Use Predicted Emotions:} Display the predicted emotions along with the corresponding bounding boxes around the detected faces on the input image. Optionally, store the predicted emotions in a data structure for further analysis or use in an application.
\end{enumerate}

\begin{figure}[h!]
    \centering
    \includegraphics[width=0.65\linewidth]{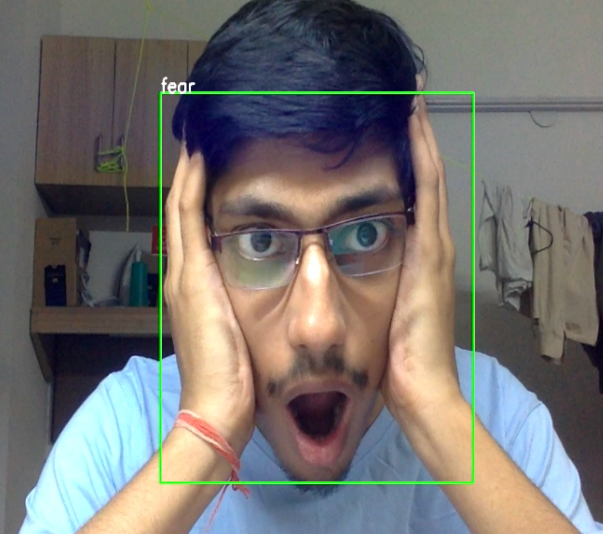}
    \caption{Predicting Fear}
    \label{fig:prototype_fear_prediction}
\end{figure}

\begin{figure}[h!]
    \centering
    \includegraphics[width=0.65\linewidth]{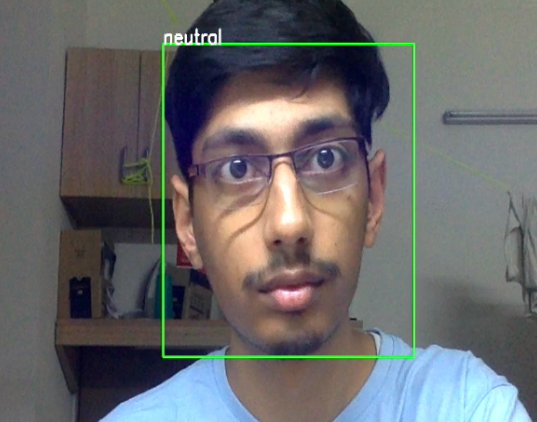}
    \caption{Predicting Neutral}
    \label{fig:prototype_neutral_prediction}
\end{figure}

\begin{figure}[h!]
    \centering
    \includegraphics[width=0.65\linewidth]{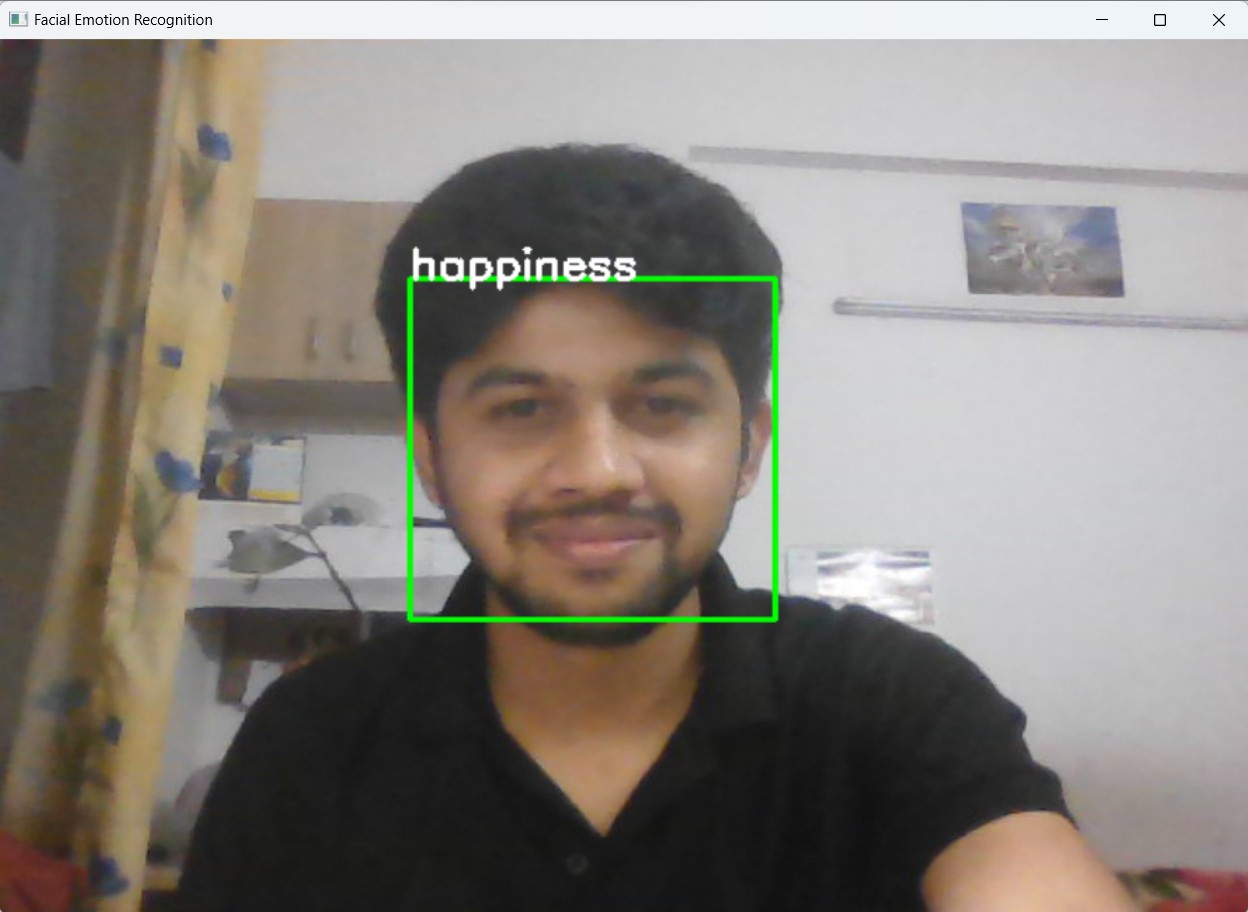}
    \caption{Predicting Happiness}
    \label{fig:prototype_happy_prediction}
\end{figure}

\begin{figure}[h!]
    \centering
    \includegraphics[width=0.65\linewidth]{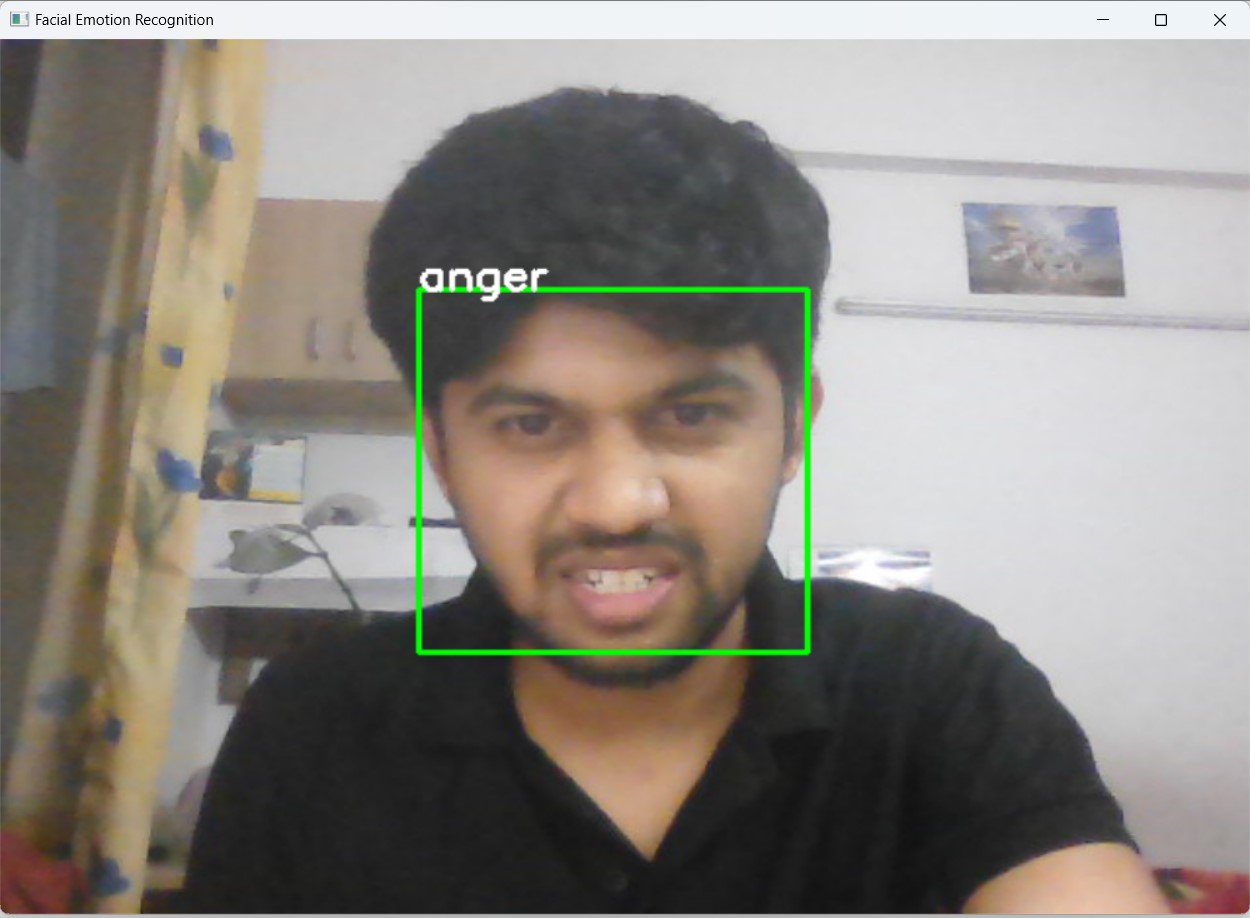}
    \caption{Predicting Anger}
    \label{fig:prototype_angry_prediction}
\end{figure}

The images - figures \ref{fig:prototype_fear_prediction}, \ref{fig:prototype_neutral_prediction}, \ref{fig:prototype_happy_prediction},\ref{fig:prototype_angry_prediction}  shown below depict the demo of our GUI prototype app performing real-time live prediction using our laptop's webcam.

\subsection{Updated GUI}
We decided to revamp the architecture a bit. We decided to leverage the power of tkinter library from python to create the GUI. When the app is loaded, it initially asks user to provide their name and also gives them two options - predict emotion from images after uploading from their system or predict emotion using their devices' webcam/camera.\par

Below - Figure \ref{fig:main_menu_app} is the main menu of our GUI app:

\begin{figure}[h]
    \centering
    \includegraphics[width=0.70\linewidth]{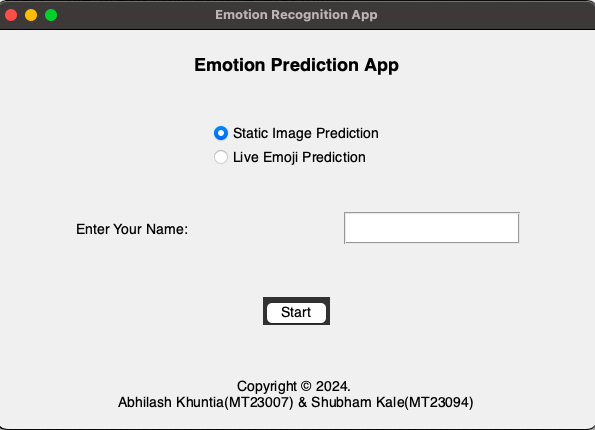}
    \caption{Main Menu of app}
    \label{fig:main_menu_app}
\end{figure}

\begin{enumerate}
    \item When opting for static image prediction, users will be prompted to grant permission for accessing their device's storage. Upon affirmative response, they will be directed to a dedicated window for static image prediction. Here, they'll be prompted to upload an image. After uploading, the GUI will exhibit the file path and present a table showcasing various emotions alongside their corresponding accuracy scores for the uploaded image. If the user selects "NO", the user would be taken back to the main menu.\par

    Below image - Figure \ref{fig:live_prediction_permission} shows the permission popup for image asking user to allow the app to access device storage:

    \begin{figure}[h!]
        \centering
        \includegraphics[width=0.70\linewidth]{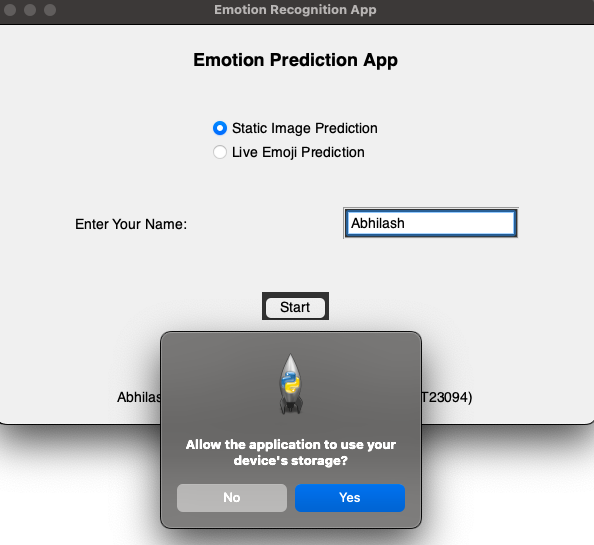}
        \caption{Image Prediction Permission}
        \label{fig:image_prediction_permission}
    \end{figure}

    Below images - Figures \ref{fig:image_prediction_nofile} and \ref{fig:image_prediction_file} show image prediction window when no file is uploaded and when file is uploaded along with the prediction.

    \begin{figure}[h!]
        \centering
        \includegraphics[width=0.70\linewidth]{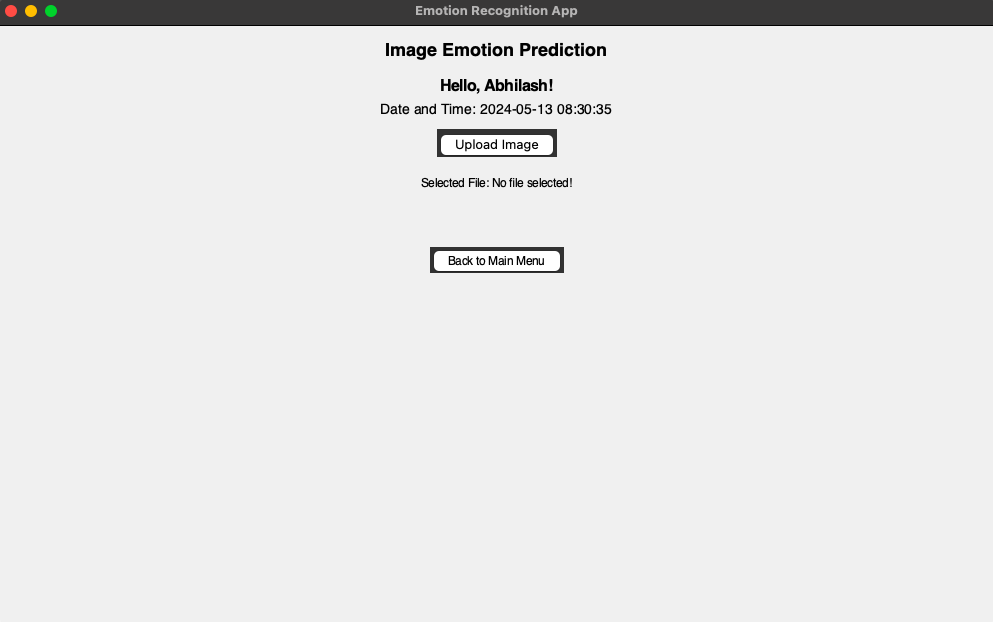}
        \caption{Image Prediction with no file uploaded}
        \label{fig:image_prediction_nofile}
    \end{figure}

    \begin{figure}[h!]
        \centering
        \includegraphics[width=0.70\linewidth]{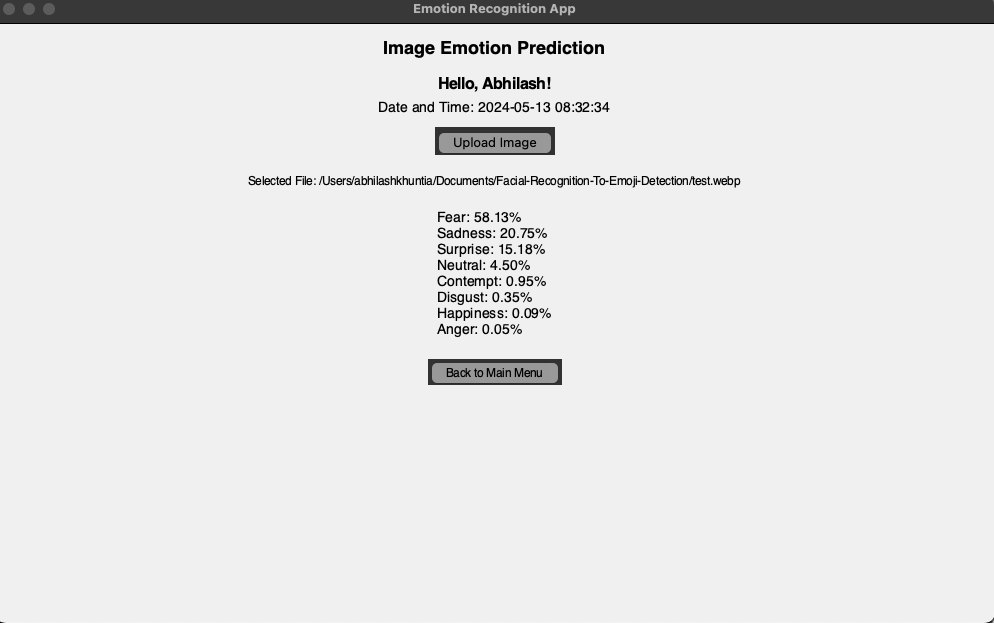}
        \caption{Image Prediction with file and its prediction}
        \label{fig:image_prediction_file}
    \end{figure}

    \item  When users opt for live prediction, they will be presented with a prompt seeking permission to access their device's webcam. If they agree by selecting "YES," a new window will unfold, featuring two distinct sections. On the left, the system will actively capture and analyze facial features from their real-time video feed. Simultaneously, the right segment will engage in predictive analysis to determine the user's emotions and subsequently display the corresponding emoji. Conversely, if users decline access by selecting "NO," they will be promptly redirected back to the main menu, ensuring their privacy preferences are respected.\par

    Below image(Figure \ref{fig:live_prediction_permission}) show the permission popup for live prediction asking user to allow the app to access device webcam/camera:

    \begin{figure}[h!]
        \centering
        \includegraphics[width=0.70\linewidth]{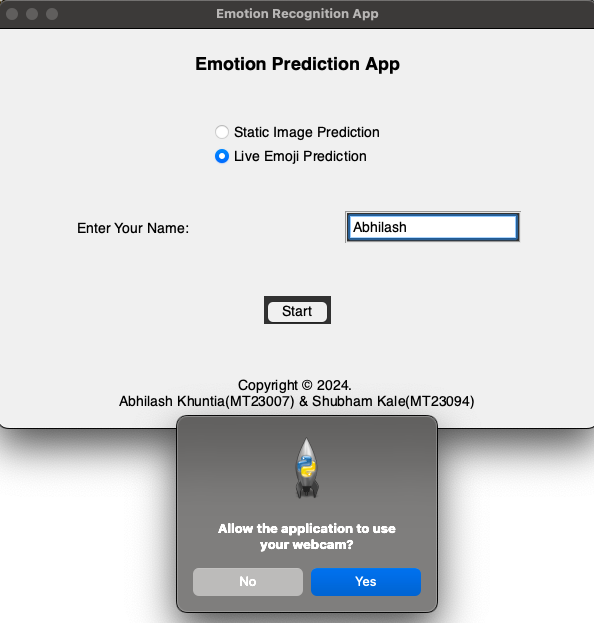}
        \caption{Live prediction permission}
        \label{fig:live_prediction_permission}
    \end{figure}

    Below image(Figure \ref{fig:live_prediction}) shows live prediction window - the window captures live video feed on the left and displays predicted emoji on the right. It also displays the name entered on main menu along with today's date:
    \begin{figure}[h!]
        \centering
         \includegraphics[width=0.70\linewidth]{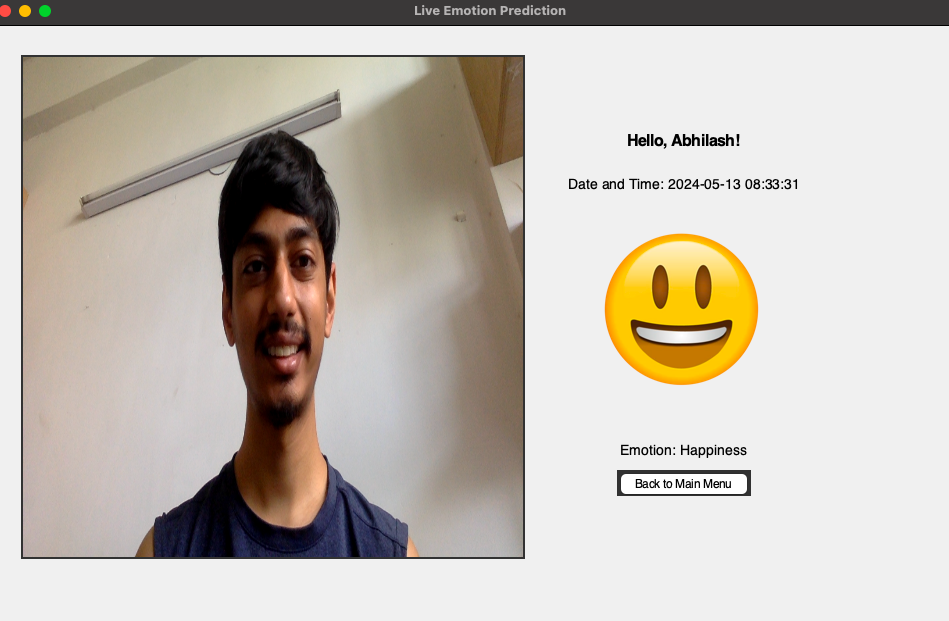}
        \caption{Live Prediction}
        \label{fig:live_prediction}
    \end{figure}
    
\end{enumerate}

\section{Proposed Student Engagement Monitoring System Using Our Model}
Our proposed architecture leverages cutting-edge facial detection technology to monitor student engagement in educational settings. The system utilizes Deep Learning ResNet Architecture to analyze facial expressions, gestures, and other visual cues in real-time. By continuously assessing students' interactions and attentiveness during class sessions, it provides valuable insights into their level of engagement.
The process begins with live video feeds captured from cameras through remote learning platforms. These video feeds are then processed by our facial detection model, which accurately identifies individual students and analyzes their facial expressions.

Based on this analysis, the system computes an engagement score for each student on a scale from 0 to 10. A score of 0 indicates minimal or no involvement, while a score of 10 signifies maximum attention and participation. Factors such as facial expressions, eye contact with the instructor, and physical gestures are taken into account to determine the overall level of engagement.

To ensure accuracy and reliability, our model undergoes rigorous training using diverse datasets encompassing various student demographics, classroom environments, and teaching styles. Additionally, continuous feedback loops enable the system to adapt and improve its performance over time.

By providing educators with real-time insight into student engagement, our architecture empowers them to tailor their teaching strategies, identify at-risk students, and optimize learning outcomes. Moreover, it fosters a more interactive and inclusive learning environment, where every student's participation is valued and encouraged.

Overall, our proposed architecture represents a significant advancement in educational technology, enabling educators to harness the power of facial detection and AI to enhance student engagement and enrich the learning experience.\par

\section{Some more applications}
\begin{enumerate}
    \raggedright
    \item \textbf{Social Media app:} While on video calls, like Apple's Facetime, we can capture live feed from the camera and display appropriate emoji. Also, now-a-days since people are engaged in sending memes to each other, we capture the expression through mobile cameras and send the reaction.\par
    
    \item \textbf{Movie Audience Reaction:} Throughout, the duration of movie, people would be monitored with delay of say about 30 min i.e say a movie is 2h:30min long, then people will be monitored for every 30 min and then at the end we find out overall reaction of the audience.\par
    
    \item  \textbf{Video Game public reaction}: Monitor the emotions/expressions of the people who are playing the video game.\par
    
    \item \textbf{Medical Field: } After a medical consultation or procedure, patients can use emoji prediction to express their emotional state. This feedback can provide insights into their overall well-being and satisfaction with the care they received. For example, if a patient selects a sad emoji, it might indicate they're experiencing discomfort or dissatisfaction, prompting further inquiry from medical staff. For patients with mood disorders or chronic illnesses affecting mental health, emoji prediction can serve as a simple yet effective tool for mood tracking. \par
    
    \item \textbf{Job interview:} In job interviews, emoji prediction can be used for assessing candidates' emotional intelligence and communication skills. During the interview process, candidates' facial expressions and body language can convey a lot about their emotional state and how they are responding to questions. By using emoji prediction, interviewers could gauge candidates' emotional reactions in real-time, providing insights into their level of engagement, confidence, and authenticity.\par
\end{enumerate}

\section{Future Work}
We can improve our model's accuracy by training it on more data i.e we could provide it with new emotions and images that contain hand gestures as well. Also, we can our model more robust and enhance accuracy in challenging conditions such as varying lighting, occlusions, or diverse demographic characteristics. Developing robust algorithms that can accurately detect and classify facial expressions across different environments and populations is essential for real-world applications.\par
As real-time applications of FER become increasingly prevalent, there is a need for efficient algorithms and hardware solutions that can perform emotion recognition tasks with low latency. Optimizing algorithms for real-time processing and developing hardware-accelerated solutions could enable FER to be seamlessly integrated into various interactive systems and applications.

\section{Conclusion}
In conclusion, our project on Emoji Prediction using Facial Detection has successfully demonstrated the integration of advanced machine learning models and real-time facial expression recognition. By leveraging CNN, LSTM, and ResNet18  approaches, we developed a robust system capable of accurately identifying facial expressions and mapping them to appropriate emojis. Our implementation, which includes a comprehensive GUI utilizing OpenCV and TensorFlow, provides an intuitive user experience for both static image and live video predictions. This project not only highlights the potential of facial detection technology in enhancing digital communication but also paves the way for future applications in diverse fields such as social media, education, healthcare, and beyond. The continuous improvement of our models and the expansion of the dataset will further enhance the accuracy and reliability of the system, making it a valuable tool for real-world applications.

\section*{References}
\bibliographystyle{plain}
\renewcommand{\bibsection}{}
\bibliography{ref.bib}

\begin{thebibliography}{10}

\bibitem{abadi2016tensorflow}
M.~Abadi et~al.
\newblock Tensorflow: A system for large-scale machine learning.
\newblock In {\em Proceedings of the 12th USENIX Symposium on Operating Systems Design and Implementation}, 2016.

\bibitem{andrew2011machine}
N.~G. Andrew.
\newblock Machine learning and ai for healthcare: Big data for better outcomes.
\newblock {\em Proceedings of the National Academy of Sciences}, 2011.

\bibitem{burges1998tutorial}
C.~J. Burges.
\newblock A tutorial on support vector machines for pattern recognition.
\newblock {\em Data Mining and Knowledge Discovery}, 2(2):121--167, 1998.

\bibitem{cohn1999recognizing}
J.~F. Cohn et~al.
\newblock Recognizing action units for facial expression analysis.
\newblock {\em IEEE Transactions on Pattern Analysis and Machine Intelligence}, 21(10):971--976, 1999.

\bibitem{dahl2012context}
G.~E. Dahl et~al.
\newblock Context-dependent pre-trained deep neural networks for large-vocabulary speech recognition.
\newblock {\em IEEE Transactions on Audio, Speech, and Language Processing}, 20(1):30--42, 2012.

\bibitem{ekman1971constants}
P.~Ekman and W.~V. Friesen.
\newblock Constants across cultures in the face and emotion.
\newblock {\em Journal of Personality and Social Psychology}, 17(2):124--129, 1971.

\bibitem{franklin2022emojigeneration}
R.~G. Franklin and Dr. A.~C. Santha~Sheela.
\newblock Emoji generation using facial emotion classifier.
\newblock {\em Mathematical Statistician and Engineering Applications}, 71(3s2):142--148, 2022.

\bibitem{goodfellow2016deep}
I.~Goodfellow et~al.
\newblock {\em Deep Learning}.
\newblock MIT Press, 2016.

\bibitem{guo2019facial}
H.~Guo and J.~Chen.
\newblock Facial expression recognition using a combination of cnn and lstm.
\newblock {\em IEEE Transactions on Affective Computing}, 2019.

\bibitem{he2016deep}
K.~He et~al.
\newblock Deep residual learning for image recognition.
\newblock In {\em Proceedings of the IEEE Conference on Computer Vision and Pattern Recognition}, 2016.

\bibitem{hochreiter1997long}
S.~Hochreiter and J.~Schmidhuber.
\newblock Long short-term memory.
\newblock {\em Neural Computation}, 9(8):1735--1780, 1997.

\bibitem{jaiswal2017automatic}
A.~Jaiswal et~al.
\newblock Automatic recognition of facial expressions using hybrid cnn and lstm models.
\newblock {\em Pattern Recognition Letters}, 123:67--73, 2017.

\bibitem{kanade2000comprehensive}
T.~Kanade et~al.
\newblock Comprehensive database for facial expression analysis.
\newblock In {\em Proceedings of the Fourth IEEE International Conference on Automatic Face and Gesture Recognition (FG'00)}, 2000.

\bibitem{kingma2014adam}
D.~P. Kingma and J.~Ba.
\newblock Adam: A method for stochastic optimization.
\newblock {\em arXiv preprint arXiv:1412.6980}, 2014.

\bibitem{ko2018brief}
B.~C. Ko.
\newblock A brief review of facial emotion recognition based on visual information.
\newblock {\em Sensors}, 18(2):401, 2018.

\bibitem{krizhevsky2012imagenet}
Alex Krizhevsky, Ilya Sutskever, and Geoffrey~E Hinton.
\newblock Imagenet classification with deep convolutional neural networks.
\newblock In {\em Advances in neural information processing systems}, pages 1097--1105. Curran Associates, Inc., 2012.

\bibitem{lecun1998gradient}
Y.~LeCun et~al.
\newblock Gradient-based learning applied to document recognition.
\newblock In {\em Proceedings of the IEEE}, 1998.

\bibitem{li2019facial}
Y.~Li et~al.
\newblock Facial expression recognition with convolutional neural networks and svm.
\newblock {\em Multimedia Tools and Applications}, 78(5):6597--6615, 2019.

\bibitem{liu2014combining}
M.~Liu et~al.
\newblock Combining deep and handcrafted features for facial expression recognition.
\newblock {\em IEEE Transactions on Image Processing}, 24(12):5644--5656, 2014.

\bibitem{mollahosseini2016affectnet}
A.~Mollahosseini et~al.
\newblock Affectnet: A database for facial expression, valence, and emotion category recognition.
\newblock {\em IEEE Transactions on Affective Computing}, 10(1):18--31, 2016.

\bibitem{papageorgiou2000trainable}
C.~Papageorgiou and T.~Poggio.
\newblock A trainable system for object detection.
\newblock {\em International Journal of Computer Vision}, 38(1):15--33, 2000.

\bibitem{pedregosa2011scikit}
F.~Pedregosa et~al.
\newblock Scikit-learn: Machine learning in python.
\newblock {\em Journal of Machine Learning Research}, 12:2825--2830, 2011.

\bibitem{ren2015faster}
S.~Ren et~al.
\newblock Faster r-cnn: Towards real-time object detection with region proposal networks.
\newblock In {\em Advances in Neural Information Processing Systems}, 2015.

\bibitem{shlens2014tutorial}
J.~Shlens.
\newblock A tutorial on principal component analysis.
\newblock {\em arXiv preprint arXiv:1404.1100}, 2014.

\bibitem{simonyan2014very}
K.~Simonyan and A.~Zisserman.
\newblock Very deep convolutional networks for large-scale image recognition.
\newblock {\em arXiv preprint arXiv:1409.1556}, 2014.

\bibitem{sutton1998reinforcement}
R.~S. Sutton and A.~G. Barto.
\newblock {\em Reinforcement Learning: An Introduction}.
\newblock MIT Press, 1998.

\bibitem{viola2001rapid}
P.~Viola and M.~Jones.
\newblock Rapid object detection using a boosted cascade of simple features.
\newblock In {\em Proceedings of the IEEE Conference on Computer Vision and Pattern Recognition}, 2001.

\bibitem{wang2019facial}
K.~Wang et~al.
\newblock Facial expression recognition based on deep learning and image processing.
\newblock {\em Neurocomputing}, 291:36--47, 2019.

\bibitem{yin20063d}
L.~Yin et~al.
\newblock A 3d facial expression database for facial behavior research.
\newblock In {\em Proceedings of the 7th International Conference on Automatic Face and Gesture Recognition (FGR06)}, 2006.

\bibitem{zhang2016affective}
Z.~Zhang et~al.
\newblock Affective computing for hci: A review.
\newblock {\em Advances in Human-Computer Interaction}, 2016.

\bibitem{zhou2020expressionnet}
Y.~Zhou et~al.
\newblock Expressionnet: A deep neural network for facial expression recognition.
\newblock {\em IEEE Transactions on Multimedia}, 2020.

\end{thebibliography}
\nocite{*}
\end{document}